\documentclass[10pt,twocolumn,letterpaper]{article}
 
\usepackage{iccv}
\usepackage{times}
\usepackage{epsfig}
\usepackage{graphicx}
\usepackage{amsmath}
\usepackage{amssymb}
 
\usepackage{algorithm}
\usepackage{algorithmic}
\usepackage{paralist}

\usepackage[pagebackref=true,breaklinks=true,letterpaper=true,colorlinks,bookmarks=false]{hyperref}

\iccvfinalcopy 
 
\def\iccvPaperID{0000000} 

\newcommand{\argmax}{\mathrm{argmax}}
\newcommand{\conv}{\mathrm{conv}}
\newcommand{\ReLU}{\mathrm{ReLU}}
\newcommand{\refeqn}[1]{(\ref{#1})}

\graphicspath{{../imgs/}}  
   
\ificcvfinal\fi
\begin{document}

\title{Maximum-Margin Structured Learning with Deep Networks for 3D Human Pose Estimation}
\author{Sijin Li\\ 
{\tt\small sijin.li@my.cityu.edu.hk}\\
\\
\\
\and
Weichen Zhang\\
{\tt\small wczhang4-c@my.cityu.edu.hk}\\ 
Department of Computer Science\\
City University of Hong Kong\\
\and
Antoni B. Chan\\
{\tt\small abchan@cityu.edu.hk}\\
\\
\\
}

\maketitle

\begin{abstract}  
This paper focuses on structured-output learning using deep neural networks for 3D human pose estimation from monocular images.
Our network takes an image and 3D pose as inputs and outputs a score value, which is high when the image-pose pair matches and low otherwise.
The network structure consists of a convolutional neural network for image feature extraction, followed by two 
sub-networks for transforming the image features and pose into a joint embedding. The score function is then the dot-product between the image and pose embeddings. The image-pose embedding and score function are jointly trained using a maximum-margin cost function.
Our proposed framework can be interpreted as a special form of structured support vector machines where the joint feature space is discriminatively learned using deep neural networks. 
We test our framework on the Human3.6m dataset and obtain state-of-the-art results compared to other recent methods.
Finally, we present visualizations of the image-pose embedding space, demonstrating the network has learned a high-level embedding of body-orientation and pose-configuration.
\end{abstract}
%
%
%
%
%
%
%
%
%
%

\vspace{-0.1in}
\section{Introduction}
Human pose estimation from images has been studies for decades. 
Due to the dependencies among joint points, it can be considered a structured-output task.
In general, human pose estimation approaches can be divided by two types: 1) prediction-based methods; 2) optimization-based methods.
The first type of approach views pose estimation as a regression or detection problem~\cite{accv2014, deeppose2014,hmlpeijcv,Jonathan2014,arjun2014iclr}. 
The goal 
is to learn the mapping from the input space (image features) to the target space (2D or 3D joint points), or to learn classifiers to detect specific body parts in the image. 
This type of method is straightforward and usually fast in the evaluation stage. 
Toshev \etal~\cite{deeppose2014} 
trained a cascaded network to refine the 2D joint locations in an image stage by stage. 
However, this approach does not explicitly consider the structured constraints of human pose.
Followup work \cite{arjun2014iclr,Jonathan2014} learned the pairwise relationship between 2D joint positions, and incorporated them into the joint predictions.
Limitations of prediction-based methods include:
the manually-designed constraints might not be able to fully capture the dependencies among the body joints; poor scalability to 3D joint estimation when the search space needs to be discretized; prediction of only a single pose when multiple poses might be valid due to partial self-occlusion.
 
Instead of estimating the target directly, the second type of approach
learns a score function, which takes both an image and a pose as inputs,  
and produces a high score for correct image-pose pairs and low scores for 
unmatched image-pose pairs.
Given an input image $x$, the estimated pose $y^*$
is the pose that
maximizes the score function, i.e., 
\begin{align} 
\vspace{-0.05in}
y^{*} =  \underset{y \in \mathcal{Y}}{\argmax} f(x, y), 
\label{eqn:structpred}
\vspace{-0.1in}
\end{align} 
where $\mathcal{Y}$ is the pose space.
If the score function can be properly normalized, then it can be interpreted as a probability distribution, 
either a conditional distribution of poses given the image, or a joint distribution over both images and joints. 
One popular model is pictorial structures~\cite{Felzenszwalb05}, where the dependencies between joints are represented by edges in a probabilistic graphical model~\cite{pgm2009}.
As an alternative to generative models, structured-output SVM~\cite{structsvm2004} 
is a discriminative method for learning a score function, %
which ensures a large margin between the score values for correct input pairs and for incorrect input pairs
\cite{labelembed2013,Catalin2009}.

As the score function takes both image and pose as input,
there are several ways to fuse the image and pose information together.
For example, the features can be extracted jointly according to the image and poses, e.g., the image features extracted around the input joint positions could be viewed as the joint feature representation of image and pose \cite{Felzenszwalb05,modec13,YiYang2011,Eichner2009BMVC}.
Alternatively, %
features from the image and pose can be extracted separately and concatenated, and the score function trained to fuse them together \cite{Ionescu2011,h36m_pami}.   
However, with these methods, the features are hand-crafted, and  performance depends largely on the quality of the features.

On the other hand, deep neural networks have been shown to be good at extracting informative high-level features \cite{overfeat2013,BengioMDR13}.
In this paper, we propose a unified framework for maximum-margin structured learning with deep neural network for human pose estimation. Our unified framework jointly learns the image and pose feature representations and the score function.
In particular, our network first extracts separate feature embeddings from the image input and from the 3D pose input.  The score function is then the dot-product between the image and pose embeddings.  The score function and feature embeddings are trained using a maximum-margin criteria, resulting in a discriminative joint-embedding of image and 3D pose.
The dot-product score function is efficient to compute, and allows for fast inference over a large set of candidate poses.
In addition, our proposed framework is quite general and can be applied to a wide range of structured-output tasks.

\section{Related work}
\label{sec:relatedwork}
Here we review recent related works in deep neural network and structured learning.

\subsection{2D pose estimation via detection with deep networks}
Traditional pictorial structure models usually apply linear filters on hand-crafted features, e.g., HoG and SIFT, to calculate the probability of the presence of body parts or adjacent body-joint pairs.
As shown in~\cite{Eichner2009BMVC}, the quality of the features are critical to the performance, and, while successful for other tasks, these hand-crafted features may not be necessarily optimal for pose estimation.
Alternatively, with sufficient data, it is %
possible to learn the features directly from training data.
In recent years, deep neural networks, especially convolutional neural networks (CNN), have been shown to be effective in learning rich features~\cite{Ali2014, Alex2012}.
Jain \etal~\cite{arjun2014iclr} trains a CNN as a sliding-window detector for each body part, 
and the resulting body-joint detection maps are smoothed using a learned pairwise relationship between joints.
Tompson \etal~\cite{Jonathan2014} extends \cite{arjun2014iclr} by feeding the body-joint detection maps into a modified convolutional layer that performs pairwise smoothing, allowing  feature extraction and pairwise relationships to be jointly optimized.
Chen \etal~\cite{Chen_NIPS14}  uses a deep CNN to predict the presence of joints and the pairwise relationships between joints, and the CNN output is then used as the input into a pictorial structure model for 2D pose estimation.
The advantage of these approaches is that the features extracted by deep networks usually lead to better performance.
However the detection-based methods for 2D pose estimation are not directly applicable to 3d pose estimation %
due to the need to discretize a large pose space -- the number of joint positions grows cubicly with the resolution of the discretization, making inference computationally expensive~\cite{Burenius2013}.
In addition, it is difficult to predict 3D coordinates from only a local window around a joint, without any other contextual information.
%


\subsection{Pose regression via deep networks}

In contrast to detection-based methods, regression-based methods aim to directly predict the coordinates of the body-joints in the image.
Toshev \etal~\cite{deeppose2014} trains a cascade CNN to predict the 2D coordinates of joints in the image, where the CNN inputs are the image patches centered at the coordinates predicted from the previous stage.
Li \etal~\cite{hmlpeijcv} use a multi-task framework to train a CNN to directly predict a 2D human pose, where auxiliary tasks consisting of body-part detection guide the feature learning.
This work was later extended for 3D pose estimation from single 2D images  \cite{accv2014}.

One disadvantage of regression-based methods is that they can only predict one pose for a given image.  This may cause difficulties on images where the pose is ambiguous due to partial-self occlusion, and hence several poses might be valid.
In contrast, our proposed model is better able to handle ambiguities since several valid image-pose pairs can have similar high scores.
 
\begin{figure*}[t]
\begin{center}  
   \includegraphics[width=0.9\linewidth]{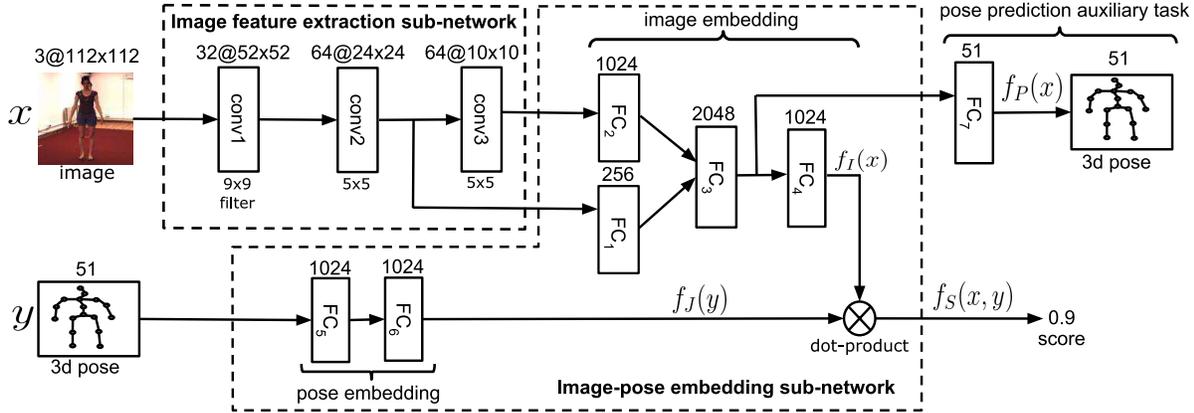}
\end{center}
\vspace{-0.1in}
   \caption{Deep-network score function.  The image input is fed through a set of convolutional layers for image feature extraction. Two separate sub-networks are used to embed the image and the pose into a common space, and the score function is the dot-product between the two embeddings.
An auxiliary 3D body-joint prediction task is used to guide the network to find good image features.
Each convolutional layer is followed by a max-pooling layer, which is not drawn to reduce clutter.}
\label{fig:imgposenet}
\vspace{-0.15in}
\end{figure*}

\subsection{Structured-output prediction and feature embedding}
Rodr{\'{\i}}guez \cite{labelembed2013}
represents the score function between word labels and images as the dot-product between the word-label feature and an image embedding, and trains a structured SVM (SSVM) to learn the weights to map the bag-of-words image features to the image embedding.
Dhungel \etal~\cite{DhungelCB14} uses structured learning and deep networks to segment mammograms.  First, a network is trained to generate a unary potential function.
Next, a linear SSVM score function is trained on the output of the deep network, as well as other potential functions. 
Osadchy \etal~\cite{Osadchy:2007} apply structured learning and CNN %
for face detection and face pose estimation. 
The CNN %
was trained to map the face image to a manually-designed face pose space. A per-sample cost function is defined with only one global minimum so that the ground-truth pose has minimum energy.
In contrast to~\cite{DhungelCB14,labelembed2013,Osadchy:2007}, we learn the feature embedding and score prediction jointly within a maximum-margin framework.

Jaderberg \etal~\cite{deepstruct2015} proposed a deep structured-output network for recognizing text in images. The score function is a conditional random field (CRF), where the input is an image and the output is a word.  The unary and higher-order potential functions of the CRF are two CNNs, which are trained to recognize single characters and n-grams in the image, and the framework is jointly trained with a maximum margin cost.
In the context of pose recognition, \cite{deepstruct2015} is a pictorial structure model with higher-order terms, whereas our method is similar to learning a non-linear embedding with a linear SSVM score function.
In particular, the main difference is that we do not manually-design the score function to encode the output structure as pairwise or higher-order terms (i.e., the CRF), but instead train the network to learn both image and pose embeddings such that a score function can be represented as dot-product.  Furthermore, the internal image representations in \cite{deepstruct2015} are strongly supervised, consisting of 
 character/n-gram classifiers, whereas the internal representations (image/pose embeddings) in our method are learned from the data.
Although both methods use a maximum-margin cost, \cite{deepstruct2015}  uses a fixed margin for all input/output pairs, whereas our method uses margin rescaling.

\vspace{-0.05in}
\subsubsection{Unsupervised joint feature embedding}

\vspace{-0.05in}
Deep networks have also been used to learn joint embeddings for multi-modal inputs. Ngiam \etal~\cite{multimodelDL2011} embed audio-video pairs by jointly training autoencoders  with a shared middle layer.
Pereira \etal~\cite{multimodelDBN2012} build a generative model for image-text pairs by adding a binary hidden layer on top of image-specific and text-specific deep Boltzmann machines.
Andrew \etal~\cite{DCCA} proposes deep canonical correlation analysis (DCCA), 
where each input view is passed through a separate deep network (implementing a non-linear transformation), 
and the networks are jointly trained so that the their outputs are maximally correlated.
In contrast to these works, our joint embedding is learned discriminatively using a maximum-margin cost.  In addition, our embedding is loosely coupled, i.e., the image and pose embeddings do not explicitly share latent variables (layers).  Rather the two embeddings are optimized through the dot-product similarity and supervised cost function, similar to learning a kernel embedding.

 
 \section{Maximum-margin structured learning}
\label{sec:structnet}
Our goal is to learn a score network that can assign maximum score to  correct image-pose pairs and low scores to other pairs.
The network structure is illustrated in Figure~\ref{fig:imgposenet}.
Our network consists of two main components: an image feature extraction sub-network and an image-pose embedding sub-network.
For the first sub-network, a CNN extracts high-level image features from the raw image.
For the second sub-network, 
the image features and pose (3D joint coordinates) are separately fed through fully-connected layers, mapping them into two embedding spaces. The score function is then the dot-product between the two embeddings.
Although the image/pose embeddings are calculated from separate sub-networks, training the full network will align the image/pose embeddings into a joint space, such that their dot-product is a suitable score function.

To train the network, we use a maximum-margin cost function that forces the score
of the ground-truth image-pose pair to be larger than other image-pose pairs by at least a margin.  We use a re-scaling margin, which is a function of the distance between the ground-truth pose and the other pose.
In order to encourage image features that preserve pose information, we add an auxiliary task consisting of 3D body-joint prediction during training.

In the following, we use $x$ to represent the image input, $y$ as the ground-truth matching pose (3D joint coordinates), $\mathcal{Y}$ as the pose space, and $\theta$ as the network parameters.

\subsection{Image feature extraction}
\vspace{-0.05in}
The goal of the image extraction sub-network is to convert the raw input image to a more compact representation with pose information preserved.
We use a deep CNN, consisting of 3 sets of convolution and max-pooling layers,  to extract image features from the image.
We use rectified linear units (ReLU)~\cite{relu2010} as the activation function in the first 2 layers, and the linear activation function in the 3rd  layer.

The outputs of the pooling layers is a set of feature maps, denoted as $\conv^j(x)$, where $j$ is the layer number. 
Each feature in the map has a receptive field in the input image, with higher layer features having larger receptive fields.
Intuitively, the higher layer features will contain global information about the pose, which would be useful for distinguishing between grossly different poses.
On the other hand, the lower layer features contain more detailed information about the pose, which will be helpful in distinguishing between similar poses.

\subsection{Image-pose embedding}
\label{sub:imagepose}
\vspace{-0.05in}
The image and pose inputs are in different spaces, and
the goal of the image-pose embedding sub-network is to project the image features and the 3D pose into a joint embedding space where they can be compared effectively.
The architecture of image and pose embedding network is shown in Figure~\ref{fig:imgposenet}. 
Inspired by \cite{Sun2014dlface,hmlpeijcv}, we use features from both the middle- and top-convolutional layers.
The middle- and top-layer features are each passed through separate fully connected layers, and then concatenated and passed through two more fully connected layers to form the image embedding $f_{I}(x)$.
Specifically,
\begin{align} 
f_{I}(x) = h_4 ( h_3 ( \begin{bmatrix} h_1(\conv^{2}(x)) \\ h_2(\conv^{3}(x)) \end{bmatrix})),
\label{equ:imgfeature}
\end{align} 
where the activation function $h_i(x) = \ReLU( W_i^T x + b_i)$ is a rectified linear unit with  weight matrix $W_i$ and bias $b_i$.

The input pose $y$ is represented by the 3D coordinates of the body-joint locations, 
the dimensions of which are strongly correlated due the dependencies among joints. 
The pose is mapped into a non-linear embedding, so that it can be more easily combined with the image embedding.
We use 2 fully connected layers for this transformation,
\begin{align}
f_{J}(y) = h_6( h_5(y)).
\end{align}

\subsection{Score prediction}
We represent the score function between the image and pose inputs $f_{S}(x,y)$ as the inner-product between the image embedding $f_{I}(x)$ and pose embedding $f_{J}(y)$, i.e., 
\begin{align}
f_{S}(x,y)  = \langle f_{I}(x), f_{J}(y) \rangle.
\end{align} 
One advantage of using inner-product is that the corresponding dimensions of the image/pose embedding vectors interact directly, which makes aligning the two embeddings easier.
Another advantage 
is that it is very efficient to calculate.
The calculation of the pose embedding does not depend on the image features, which means it can be calculated offline if the set of candidate poses is fixed.

Training the network will map the image and pose into similar embedding spaces, where their dot-product similarity serves as a suitable score function. This can be loosely interpreted as learning a multi-view ``kernel'' function, where the ``high-dimensional'' feature space is the learned joint embedding.

Our score function can also be interpreted as a %
SSVM, where the joint features
are the element-wise product between the learned image and pose embeddings, 
\begin{equation}
  f'_{S}(x,y) = \langle w, f_{I}(x) \circ f_{J}(y) \rangle
\end{equation}
where $\circ$ indicates element-wise multiplication, and  $w$ is the SSVM weight vector.  
The equivalence is seen by noting that during network training the  weights $w$ can be absorbed into the embedding functions $\{f_I, f_J\}$.
In our framework, these embedding functions are discriminatively trained.

\begin{figure*}[t]
\begin{center}     
\begin{tabular}{c|c}
{\footnotesize \bf network structure for finding the most-violated pose}
&
{\footnotesize \bf network structure for maximum-margin training}
\\
   \includegraphics[width=0.49\linewidth]{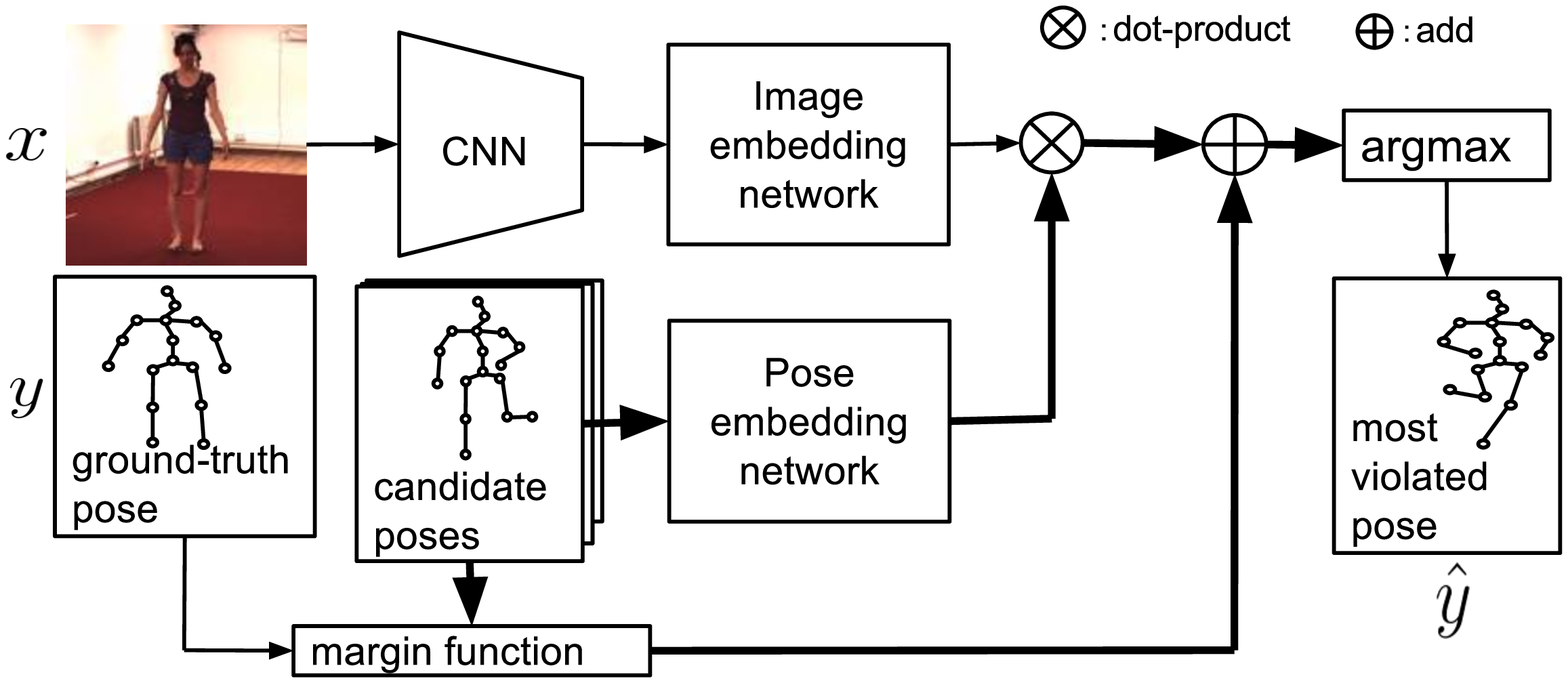}
   &
  \includegraphics[width=0.48\linewidth]{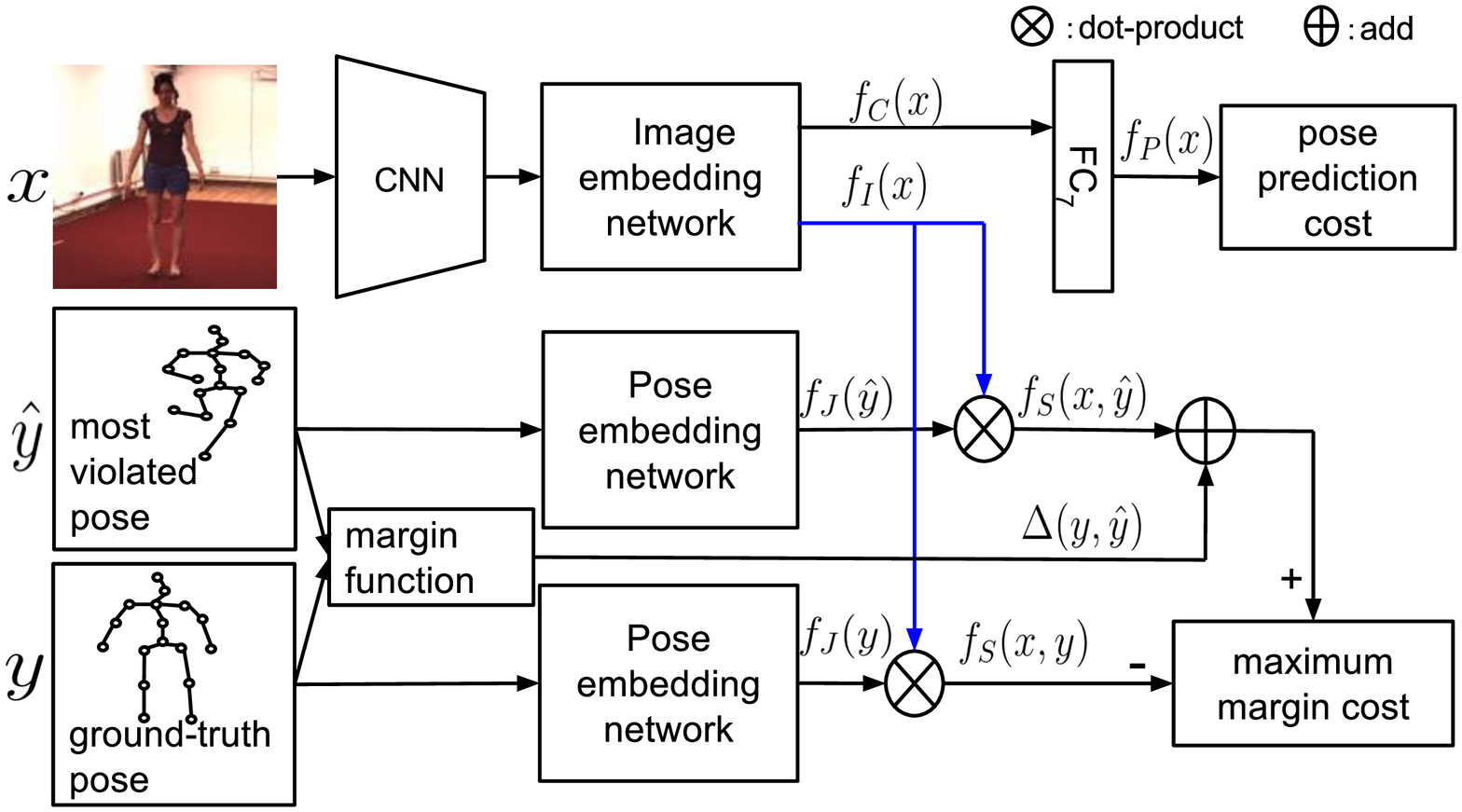}
\end{tabular} 
\end{center}  
\vspace{-0.15in}
   \caption{(left) Network structure for calculating the most violated pose.  For a given image, the score values are predicted for a set of candidate poses.  The re-scaling margin values are added, and the largest value is selected as the most-violated pose. Thick arrows represent an array of outputs, with each entry corresponding to one candidate pose.
 (right) Network structure for maximum-margin training. Given the most-violated pose, the margin cost and pose prediction cost are calculated, and the gradients are passed back through the network.}
\label{fig:trainnet}
\vspace{-0.15in}
\end{figure*}

\subsection{Maximum margin cost}
\vspace{-0.05in}
Inspired by maximum-margin structured SVM~\cite{ssvm2005}, we use a maximum margin cost to learn the score function.
The maximum margin cost ensures that the difference between the scores of two input pairs is at least a particular value (i.e., the margin). Different from the standard SVMs, with structured-SVM can have a margin that changes values based on dissimilarity between the two input pairs.

Similar to the structured-SVM, 
we use the margin re-scaling surrogate loss,
\begin{eqnarray}
 \mathcal{L}_{M}(x,y, \hat{y}) = \max(0, f_{S}(x,\hat{y}) + \Delta(\hat{y}, y) - f_{S}(x,y)), 
\label{equ:mmcost} 
\end{eqnarray}
where $(x,y)$ is a training image-pose pair, 
$\Delta(y,y')$ is a non-negative margin function between two poses, and
$\hat{y}$ is the pose that most violates the margin constraint\footnote{Note that $\hat{y}$ depends on the input $(x,y)$ and network parameters $\theta$. To reduce clutter, we write $\hat{y}$ instead of $\hat{y}(x,y,\theta)$ when no confusion arises.},
\begin{align}
\hat{y} &= \underset{y'\in \mathcal{Y}}{\argmax} f_{S}(x,y') + \Delta(y,y') - f_{S}(x,y).
\label{equ:mvpose}
\end{align}

 Intuitively, a pose with a high predicted score, but that is far from the ground-truth pose, is more likely to be the most violated pose. 
For the margin function, we use the mean per joint error (MPJPE), i.e., 
\begin{align}
\Delta(y,y') = \frac{1}{J}\sum_{j=1}^{J} \| y_{j} - y'_{j}\|,
\end{align}
where $y_{j}$ indicates the 3D coordinates of $j$-th joint in pose $y$, and $J$ is the number of body-joints. 

When the loss function in \refeqn{equ:mmcost} is zero, then the score of the ground-truth image-pose pair $(x,y)$ is at least larger than the margin for all other image-pose pairs $(x,y')$, 
	\begin{align}
	f_{S}(x,y) \geq f_{S}(x,y') + \Delta(y', y), \forall y'\in{\mathcal Y}.
	\end{align}
On the other hand, if  \refeqn{equ:mmcost} is greater than 0, 
then there exists at least one pose $y'$ whose score $f(x,y')$ violates the margin.

\subsection{Multi-task global cost function}
\vspace{-0.05in}
Following \cite{accv2014, hmlpeijcv},  in order to encourage the image embedding to preserve more pose information, we include an auxiliary training task of predicting the 3D pose.
Specifically, we add a 3D pose prediction layer after the penultimate layer of the image embedding network,
\begin{align}
f_{P}(x) = g_7(h_3),
\end{align}
where $h_3$ is the output of the penultimate layer of the image embedding, and $g_i(x)=\tanh(W_i^Tx+b_i)$ is the $\tanh$ activation function.
The cost function for the pose prediction task is the square difference between the ground-truth pose and predicted pose, 
\begin{align}
\label{eqn:costpredictpose}
\mathcal{L}_{P}(x, y) = \| f_{P}(x) - y\|^{2}.
\end{align}
Finally, given a training set of image-pose pairs $\{(x^{(i)}, y^{(i)})\}_{i=1}^N$,  our global cost function consists the structured maximum-margin cost,  pose estimation cost, as well as a regularization term on the weight matrices,
\begin{equation}
\begin{split}
\text{cost}(\theta) &= \frac{1}{N} \sum_{i=1}^{N} \mathcal{L}_{M}(x^{(i)},y^{(i)}, \hat{y}^{(i)}) 
\\
& + \frac{1}{N}\lambda\sum_{i=1}^{N} \mathcal{L}_{P}(x^{(i)}, y^{(i)}) 
+ \alpha\sum_{j=1}^7 \|W_j\|^2_F
\end{split}
\end{equation} 
where $i$ is the index for training samples,  $\lambda$ is the weighting for pose prediction error, $\alpha$ is the regularization parameter, and $\theta=\{(W_i,b_i)\}_{i=1}^7$ are the network parameters.
Note that gradients from $\mathcal{L}_{P}$ only affect the CNN and high-level image features ($\text{FC}_{1}$-$\text{FC}_{3}$), and have no direct effect on the pose embedding network or image embedding layer ($\text{FC}_{4}$). Therefore, we can view the pose prediction cost as a regularization term for the image features.
Figure \ref{fig:trainnet} shows the overall network structure for calculating the max-margin cost function, as well as finding the most violated pose.

\section{Training Algorithm} 
\label{sec:Training}
\vspace{-0.1in}

We use back-propagation~\cite{Rumelhart:1988} with stochastic gradient descent (SGD)  to train the network.
Similar to SSVM~\cite{Joachims:2009}, 
our training procedure iterates between finding the most-violated poses and updating the network parameters:
\begin{compactenum}
\item Find the most-violated pose $\hat{y}$ for each training pair $(x,y)$ using the pose selection network  with current network parameters (Fig.~\ref{fig:trainnet} left);
\item Input $(x,y,\hat{y})$
into the max-margin training network (Fig.~\ref{fig:trainnet} right) and run back-prop to update parameters.
\end{compactenum}
We call the tuple $(x,y,\hat{y})$ the extended training data. 
The training data is processed in mini-batches.
We found that using momentum between mini-batches,
which updates the parameters  using the weighted average of the current gradient and previous update, 
always hinders convergence.
This is because the maximum-margin cost selects different most-violated poses in each batch, 
which makes the gradient direction change rapidly between batches.
To speed up the convergence of SGD, we use a line-search to find the best step-size for each mini-batch update.
This was necessary because the the back-propagated gradients have high dynamic range, which stems from the cost function consisting of the difference between network outputs.

Although our score calculation is efficient, it is still computationally expensive to search the whole pose space to find the most-violated pose. 
Instead, we form a candidate set $\mathcal{Y}_{\mathcal{B}}$ for each mini-batch, and find the most-violated poses within the candidate set. 
The candidate set consists of $C$ poses sampled from the pose space $\mathcal{Y}$.  
In addition, we observed that some poses are %
selected as the most-violated poses multiple times during training.
Therefore, we also maintain a working set of most-violated poses, and include the top $K$ most-frequent violated poses in the candidate set.

Our training procedure is summarized in Algorithm~\ref{algo:snlearn}.
Note that the selection of the most-violated pose from a candidate set, along with the back-propagation of the gradient for that pose, can be interpreted as a max-pooling operation over the candidate set.

\begin{algorithm}[tbh]
\begin{algorithmic}
 \STATE {\bf input}:   training set $\{(x^{(i)},y^{(i)})\}_{i=1}^N$, pose space $\mathcal{Y}$, number of iterations $M$, number of mini-batches $B$, number of candidate poses $C$, number of most frequent violated poses $K$.
 \STATE {\bf output}:  network parameters $\theta$.
 \STATE $\mathcal{V}= \emptyset$ \quad\COMMENT{working set of most-violated poses}
\FOR[loop over the whole training set]{$t=1\,\TO \,M$}
 \FOR[loop over mini-batches]{$b=1\,\TO \,B$}
\STATE $\mathcal{B}$ = ReadBatch() 
\STATE \COMMENT{get the current set of candidate poses $\mathcal{Y}_{\mathcal{B}}$}
\STATE $\mathcal{Y}_{\mathcal{B}} = \text{UniformSample}(\mathcal{Y}, C)$ \COMMENT{get $C$  poses}
\STATE $\mathcal{Y}_{\mathcal{B}} = \mathcal{Y}_{\mathcal{B}} \cup \text{KMostFrequent}(\mathcal{V},K)$  
\STATE \COMMENT{build the extended training data $\mathcal{D}$}
\STATE $\mathcal{D}=\emptyset$
\FORALL {$ (x,y) \in \mathcal{B}$}
\STATE \COMMENT{calculate the most violated pose for $(x,y)$}
\STATE $\hat{y} = \underset{y'\in \mathcal{Y}_{\mathcal{B}}}{\argmax} \langle f_{I}(x), f_{J}(y') \rangle + \Delta(y,y')$
\STATE $\mathcal{D} = \mathcal{D}  \cup   (x,y,\hat{y})$ \COMMENT{add to extended data}
\STATE $\mathcal{V} = \mathcal{V}  \cup  \hat{y}$  \COMMENT{add to working set of violated poses}
\ENDFOR
\STATE \COMMENT{update network parameters}
\STATE $\text{StepSize}=\text{LineSearch}(\text{cost}, \mathcal{D},\theta)$
\STATE $\theta = \text{SGD}(\text{cost}, \mathcal{D}, \theta, \text{StepSize})$
       \ENDFOR
\ENDFOR
  \caption{Max-margin structured-network training}
  \label{algo:snlearn}
\end{algorithmic}

\end{algorithm}
\vspace{-0.2in}
\section{Experiments} 
\label{sec:Experiments}
\vspace{-0.1in}
In this section, we evaluate our maximum margin structured learning network on human pose estimation dataset.
\vspace{-0.3in}
\subsection{Dataset}
\vspace{-0.05in}
We evaluate on the Human3.6M dataset~\cite{h36m_pami}, which contains around 3.6 million frames of video. 
The videos are recorded with four RGB camera, along with a MoCap system for measuring the joint positions.
We treat the four RGB images separately, and project the MoCap coordinates to each camera coordinate system as the ground-truth pose. 

As in \cite{h36m_pami,accv2014}, the image input is a cropped image around the human.
The training images are obtained by extracting a square image according to the bounding box provided in Human3.6M dataset~\cite{h36m_pami}, and resizing it to 128$\times$128.
As in~\cite{Alex2012}, we augment the image training set by local translations and by adding random pixel noise during training.   
For local translations, a 112$\times$112 sub-image is randomly selected from the training image.
For pixel noise, random noise is added to all pixels according to the RGB covariance matrix calculated over the whole training set.
The 3D pose input is a vector of the 3D coordinates of 17 body-joints.

\vspace{-0.15in}
\subsection{Experiment setup}
\vspace{-0.12in}
We follow the same protocol as in~\cite{accv2014} for the training and test set -- we use 5 subjects (S1, S5, S6, S7, S8) for training and validation, and 2 subjects (S9, S11) for testing.

Our structured-output network (denoted as StructNet) is trained using the algorithm from Section \ref{sec:Training}.
Given a test image, ideally, the predicted pose should be found by searching the entire pose space $\mathcal{Y}$ for the pose with maximum score, as in \refeqn{eqn:structpred}.
However, the pose space is continuous and exhaustive search is computationally intractable.
Instead, we consider several approaches to approximate the search:

\begin{compactitem}
\item StructNet-Max -- the predicted pose is the pose in the training set with maximum score.
\item StructNet-Avg($A$) --  since the training and test sets contain different subjects, the poses in the training 
set will not perfectly match the subjects in the test set. To allow for more pose variation, the predicted pose is the average of the $A$ training poses with highest scores.
\item StructNet-Avg($A$)-APF -- the problem with using StructNet-Avg is that the average pose is not guaranteed to be a valid pose.  We use the annealing particle filtering (APF)~\cite{deutscher2005articulated} to generate a valid pose that best matches the pose estimated with StructNet-Avg($A$).  
Specifically, APF adjusts the joint angles of a template pose to minimize the MPJPE with the StructNet-Avg pose.  The template pose, which is a neutral ``T'' pose from the test subject, 
is initialized with the joint-angles from one of the top $A$ poses.  After APF converges, the joint-angles 
are converted into 3D joint coordinates.
\end{compactitem}
The pose estimates on the test set are evaluated using MPJPE~\cite{h36m_pami}.
We also compare against multi-task deep networks (DconvMP-HML) \cite{accv2014}, which trains a CNN using  the pose prediction cost (Eq.~\ref{eqn:costpredictpose}), and LinKDE, the best performing method in \cite{h36m_pami}.

\subsection{Implementation details}
\vspace{-0.1in}
The sizes of the network layers are shown in Figure \ref{fig:imgposenet}.
We follow the multi-task framework in~\cite{accv2014} to initialize the weights for the convolutional layers. All the weight matrices for other layers are randomly initialized.
When training the maximum-margin network, we fix the weights in the convolutional layers while still doing the  data augmentation of the input image.
The line-search was performed over the range $[10^{-7},10^{2}]$.
We approximate the pose space $\mathcal{Y}$ with all the poses in the training set. The batch size is 128, 
and the size of the sampled candidate set is $C=2000$. The number of most-frequent violated poses is $K=10$.
The weight for the auxiliary prediction task is $\lambda=1$, and the regularization parameter is $\alpha=0.0001$.
We use dropout in the fully-connected layers $\{h_1, h_{2}\}$. 
The dropout rate is 75\%. Our network is implemented in Theano \cite{Bastien-Theano-2012}.

\subsection{Experiment results}
\vspace{-0.05in}

\begin{figure}
\begin{center}  
\includegraphics[width=0.75\linewidth]{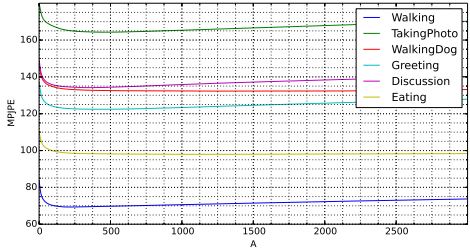}
\end{center}
\vspace{-0.2in}
   \caption{Pose error when averaging the top-$A$ highest scoring training poses (StructNet-Avg($A$)). $A=500$ represents $\sim$0.5\% of the poses in the training set.
}
\label{fig:topkvsmpjpe}
\vspace{-0.2in}
\end{figure}

Table~\ref{tab:hmlpeeval} presents the MPJPE results on the test set for each action, as well as the overall average.
We first compare the different methods for estimating the pose from StructNet.
On all actions, StructNet-Avg (the average of the top scoring poses) yields better results than StructNet-Max (the maximum scoring pose), with overall reduction in error of about 10\% when $A=500$.
Figure~\ref{fig:topkvsmpjpe} plots the error versus different values of $A$.  The error stabilizes between $A=500$ and $A=1000$, which represents $\sim$0.5\% of the poses in the training set.
Furthermore, applying APF to the average pose from StructNet-Avg yields a valid pose with roughly the same MPJPE as StructNet-Avg.

Comparing to previous works, the error for StructNet-Avg is less than
DconvMP-HML \cite{accv2014} and LinKDE \cite{h36m_pami} on all actions.  The overall error is reduced 9.2\% (from 133.54 for DconvMP-HML to 121.31 for StructNet-Avg(500)-APF).
Also note that our method generates valid poses, whereas DconvMP-HML and LinKDE do not.

Next, we consider the role of the auxiliary pose prediction task in our network.  
We evaluate the performance of the auxiliary pose prediction on the test set (denoted as StructNet-Pred in Table~\ref{tab:hmlpeeval}).    
Overall, the performance of the auxiliary pose prediction task is similar to that of \cite{accv2014},
which also uses a CNN for 3D pose estimation, but inferior to the poses obtained using the score function. 
We also test the effect of the auxiliary task on training the network.
When removing the auxiliary task, i.e., $\lambda=0$, the pose error increases  (denoted as StructNet$^*$-Max in Table~\ref{tab:hmlpeeval}).  This demonstrates that the auxiliary task helps the network to converge to a good local optimum.

To justify the design choice of our pose embedding sub-network, we trained the whole network with different forms of pose embeddings: raw 3D joint coordinates, 1-layer network with fixed random weights, 1-layer network, and 2-layer network.  The results are presented in Table~\ref{tab:cmpposeembedding}. The network using no embedding (raw joint coordinates) has the highest error, while the 2-layer pose embedding has the lowest error, which suggests that embedding the pose in a suitable high-dimensional space is necessary.

Finally, to demonstrate robustness of our framework, we trained a network for each action category in Human3.6m (using the same network parameters), and evaluated on the online hidden test set\footnote{The action ``Direction'' is not included due to video corruption.}.
The results are presented in Table~\ref{tab:h36meval}. On average, the proposed framework achieves 8.8\% lower error than LinKDE~\cite{h36m_pami}.

\begin{table*}[tbh]
\begin{center} 
\resizebox{15.5cm}{!}{
\begin{tabular}{|l|c|c|c|c|c|c|c|}
\hline  
Action & Walking & Discussion & Eating & Taking Photo & Walking Dog & Greeting & All \\
\hline\hline 
LinKDE(BS)~\cite{h36m_pami}& 97.07 (37.14)     &183.09 (116.74)          & 132.50 (72.53)  & 206.45 (112.61)&177.84 (122.65)     & 162.27 (88.43) & 162.25 (104.43)\\
DconvMP-HML~\cite{accv2014} & 77.60 (23.54)     &148.79 (100.49)          & 104.01 (39.20)  & 189.08 (93.99)&146.59 (75.38)       & 127.17 (51.10) & 133.54 (81.31)  \\
\hline
StructNet-Max & 83.64 (27.44)& 149.09 (108.93) & 109.93 (51.28) & 179.92 (93.50) & 147.24 (85.62) & 136.90 (64.71) & 135.63 (86.60)\\
StructNet-Avg(20)&75.01 (25.60)& 140.90 (110.07) & 104.10 (51.39)& 173.26 (93.71)&139.47 (86.67) & 129.08 (65.11) & 128.11 (87.18)\\ 
StructNet-Avg(500)& 69.75 (21.42)&134.37 (110.04) &98.19 (49.49)&164.28 (90.60) & 132.53 (85.91) & 122.44 (61.83) & 121.46 (85.65)\\
StructNet-Avg(500)-APF & 68.51 (22.21)&134.13 (112.87) &97.37 (51.12)&166.15 (92.95) & 132.51 (87.37) & 122.33 (64.56) & 121.31 (87.95) \\
StructNet-Avg(1500)       & 71.46 (19.75)& 137.18 (110.91)&98.01 (47.20)&166.62 (88.89)& 132.26 (83.34)& 124.58 (60.64) & 123.04 (85.17)\\
StructNet-Avg(1500)-APF & 69.97 (20.66)& 136.88 (113.93)&96.94 (49.03)&168.68 (91.55)& 132.17 (85.37)& 124.74 (63.92) & 122.85 (87.77)\\
\hline
StructNet-Pred &  84.85 (24.17) & 148.82 (102.63)& 121.57 (50.47) & 179.39 (83.72)& 151.92 (76.26) & 133.79 (56.16) & 133.79 (56.16)\\
StructNet$^{*}$-Max &87.15 (32.01)&161.62 (121.27)& 119.50 (73.04) &  196.24 (106.39) & 154.91 (99.30) & 145.30 (76.80) & 145.74 (99.69) \\ 
\hline   
\end{tabular}
}  
\end{center}
\vspace{-0.1in}
\caption{Results on Human3.6m: 
the MPJPE on the test set is calculated in millimeters (mm), with standard deviation parentheses.
} 
\label{tab:hmlpeeval}
\vspace{-0.1in}
\end{table*}

\begin{table*}[tbh]
\begin{center} 
\resizebox{17.5cm}{!}{
\begin{tabular}{|l|c|c|c|c|c|c|c|c|c|c|c|c|c|c|c|}
\hline  
Action & Discussion & Eating & Greeting & Phoning & Posing & Purchase & Sitting & SittingDown & Smoking & TakingPhoto & Waiting & Walking & WalkingDog & WalkingTogether & Avg \\
\hline\hline 
LinKDE(BS)~\cite{h36m_pami}& 108   &91    &129   &104   &130     &134    & 135    & 200    & 117   & 195    & 132     &115   & 162    & 156    & 133.81 \\
DconvMP-HML~\cite{accv2014} &103.11 &91.68 & 108.38&109.49 & 116.45 &145.24 &  145.14& 329.96 & 110.35 & 174.97 &  112.43 &99.16 & 153.29 & 116.44 &136.47\\
\hline
StructNet-Avg(500)  & 92.97 &76.70 &98.16 &92.70   & 106.86 &140.94 & 135.46 & 260.75 & 98.03  & 170.83 &  105.11 &99.40 & 138.53 & 109.30 & 122.03\\
StructNet-Avg(500)-APF&92.74& 76.38& 98.45& 92.73& 107.22& 141.21& 136.32& 265.39& 97.95& 171.71& 105.16& 99.44& 139.21& 110.28 & 122.62\\
\hline  
\end{tabular}%
}  
\end{center}
\vspace{-0.1in}
\caption{Experimental results on the online (hidden) test set of Human3.6m.} 
\label{tab:h36meval}
\vspace{-0.1in}
\end{table*}

\begin{table}
\begin{center} 
\small
\begin{tabular}{lc}
\hline  
Embedding (dim.)  &  Walking \\
\hline 
raw pose (51)  & 114.24 (45.62) \\
1-layer, random weights (1024) & 87.94 (35.61) \\
1-layer (1024)   & 86.48 (32.26) \\
2-layer (1024) & 83.64 (27.44)  \\
\hline
\end{tabular}
\end{center}
\vspace{-0.1in}
\caption{Comparison of different methods for pose embeddings.}
\label{tab:cmpposeembedding}
\vspace{-0.2in}
\end{table}

\begin{figure*}[th!]
\begin{center}  
\begin{tabular}[t]{cc}
\raisebox{2.0in}{\footnotesize (a)}
\includegraphics[width=0.32\linewidth]{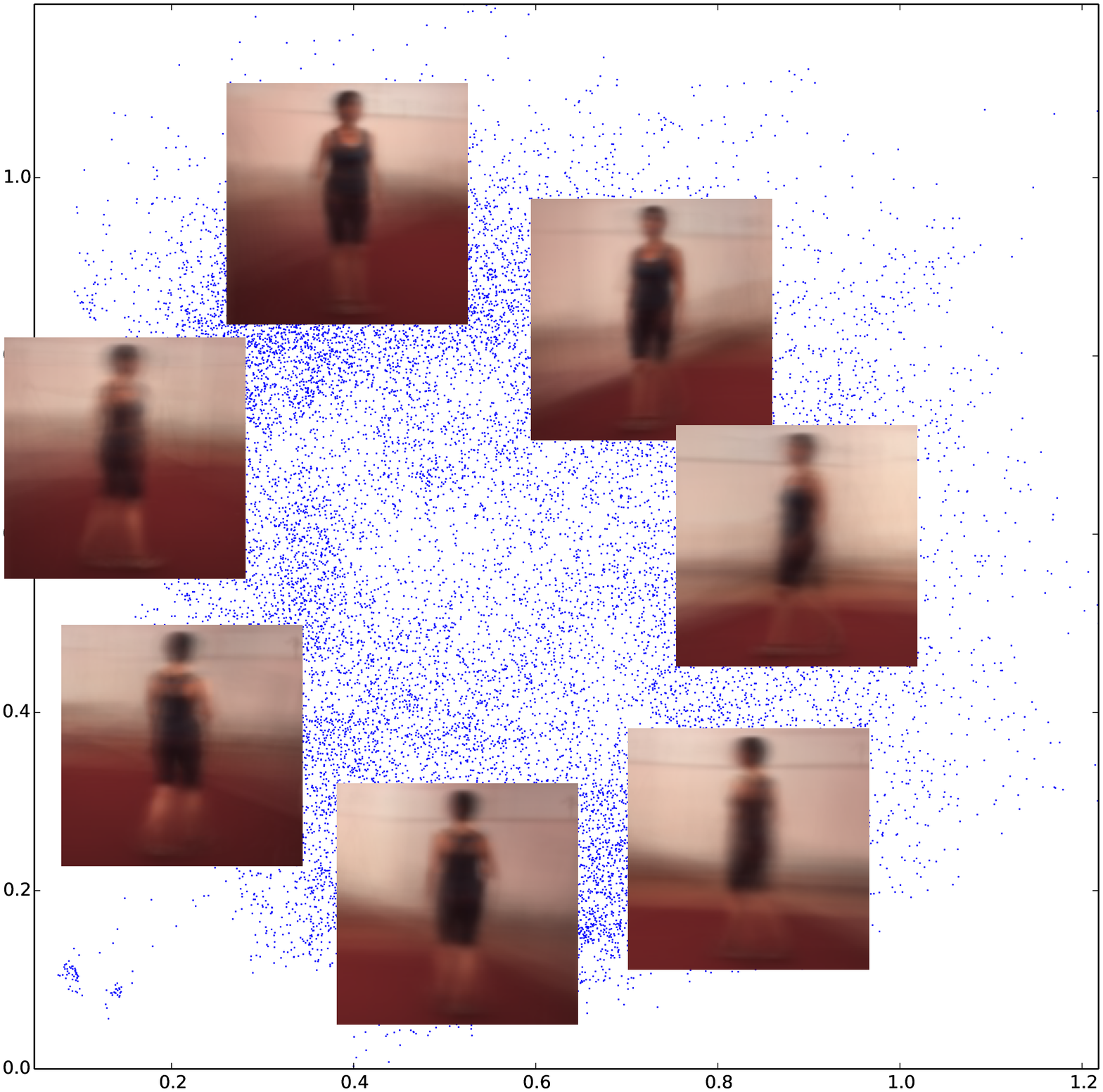} 
&      
\raisebox{2.0in}{\footnotesize (b)}
\includegraphics[width=0.32\linewidth]{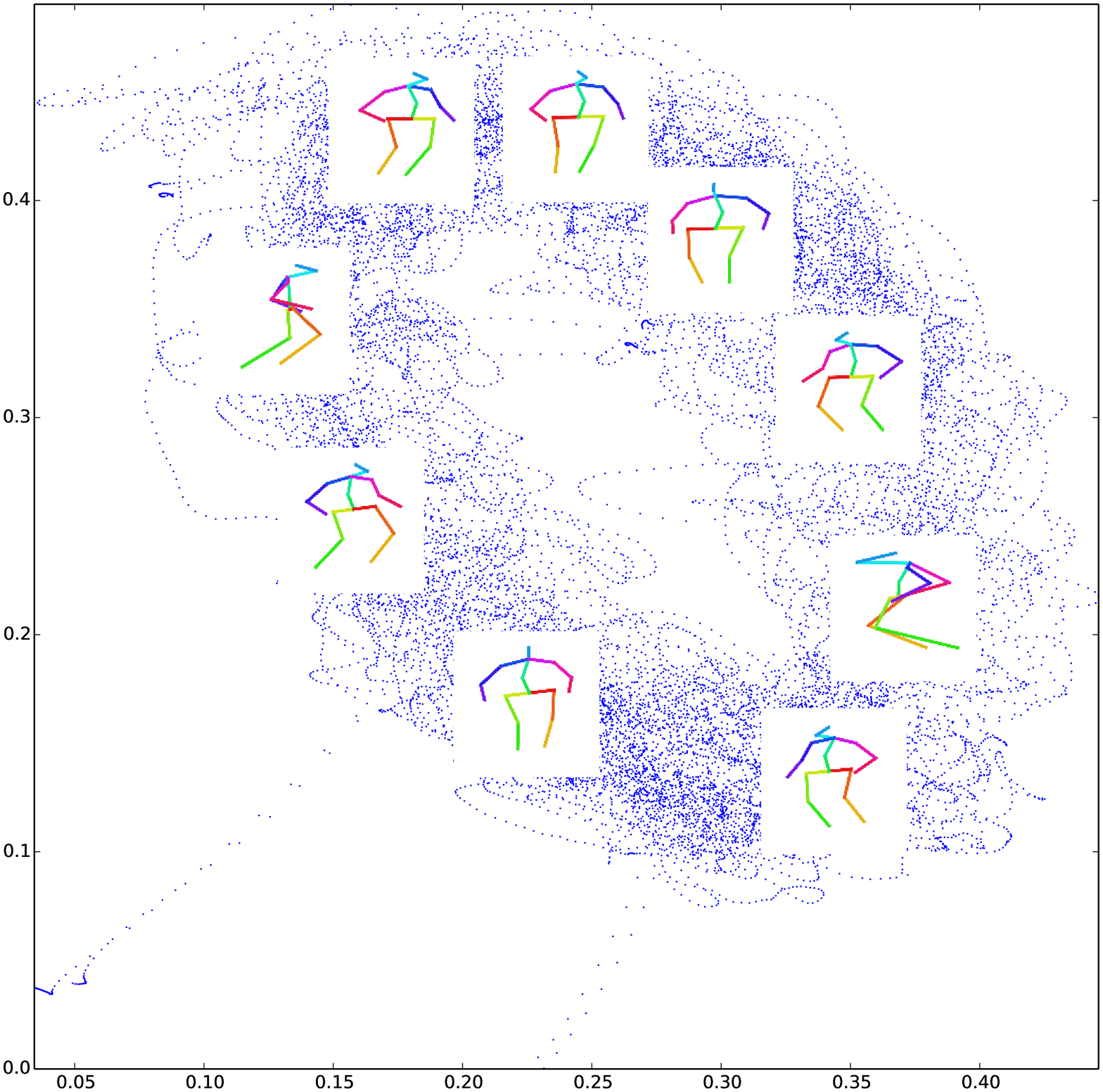}
\\   
\raisebox{1.2in}{\footnotesize (c)}
\includegraphics[width=0.33\linewidth]{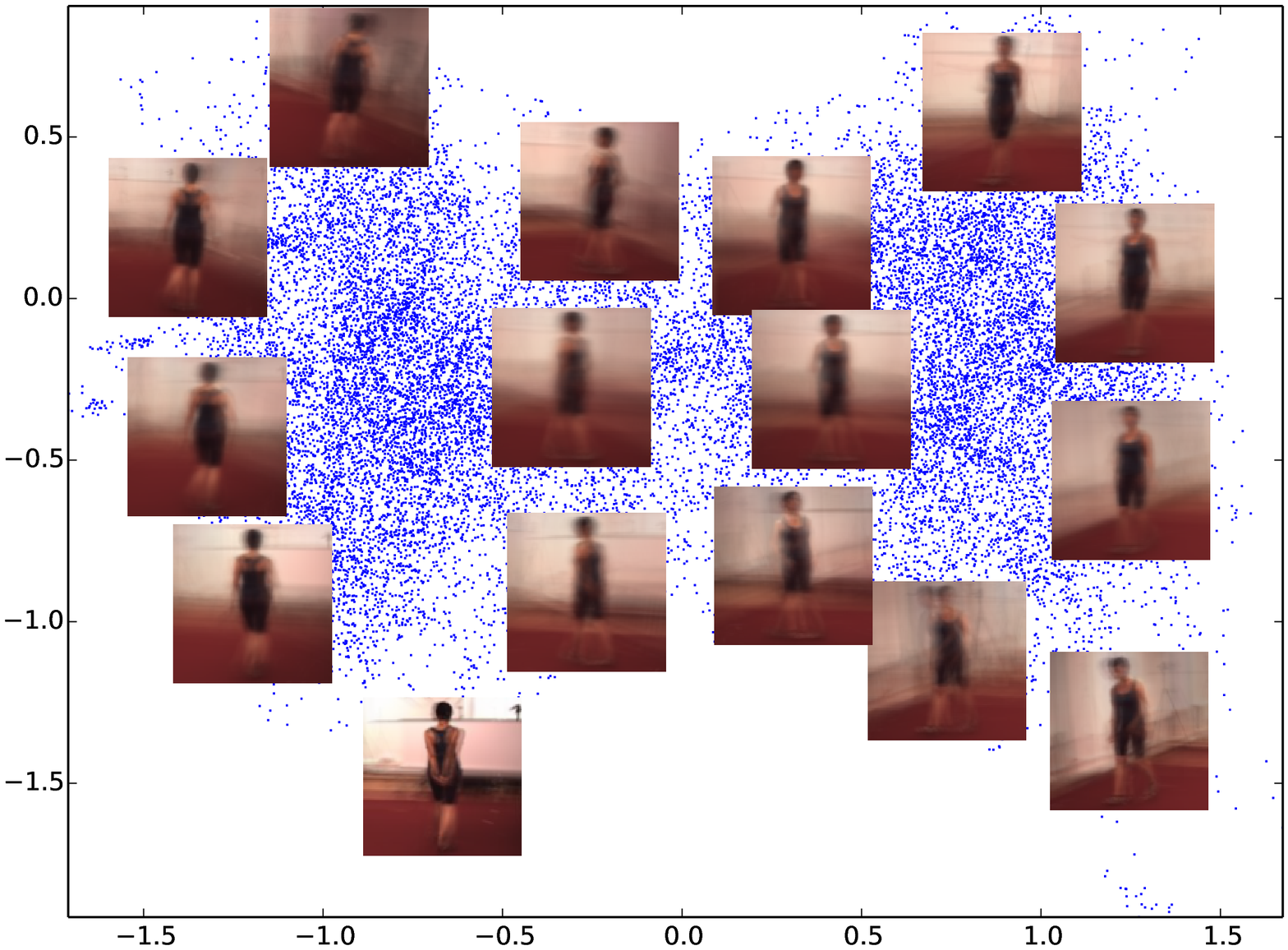}
& 
\raisebox{1.2in}{\footnotesize (d)}
\includegraphics[width=0.33\linewidth]{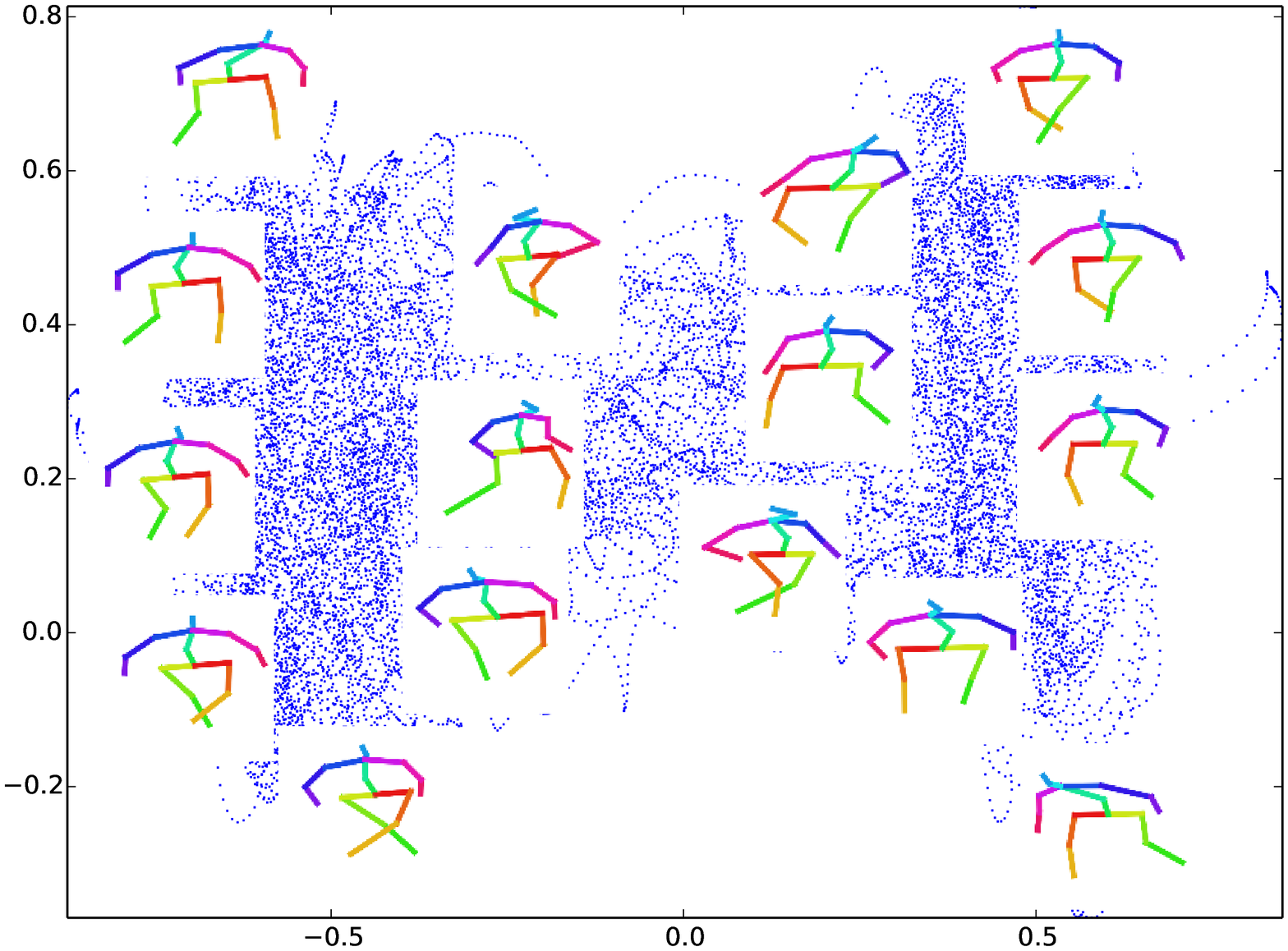}  
\end{tabular}
\end{center} 
\vspace{-0.2in}
   \caption{
   Visualizations of the learned image-pose embedding: (a) visualization of the two highest-variance features in the learned image embedding, and (b) the corresponding features in the pose embedding.
   (c) visualization of the two PCA coefficients of the learned image embedding and (d) pose embedding.
   In the pose plots, red/orange correspond to the right arm/leg, purple/green to the left arm/leg, and the cyan ``nose'' points in the forward direction of the person.}
\label{fig:featvis} 
\vspace{-0.2in}
\end{figure*}

\vspace{-0.1in}
\section{Visualization of image-pose embedding}
\label{sec:featvis}
\vspace{-0.1in}

In this section we visualize the latent features learned in the image-pose embedding.
We first look at the 2 feature dimensions of the image embedding with the highest variance  over all the training images.
Figure~\ref{fig:featvis}a plots the values of these 2 features for each of the training images.
To visualize the meaning of the features, in each local region, we show the average of the input images\footnote{For better visualization, we only use the images from a single subject.} corresponding to the feature points in that region.
Figure~\ref{fig:featvis}b shows a similar plot for the same 2 feature dimensions in the pose embedding, with average poses over local regions of the space.  
The top-2 features in the embedding correspond to the orientation of the person.
For example, in Figure~\ref{fig:featvis}a, the average image in the upper-part of the plot is a frontal view of the person, while the average image in the lower-part is the back view (similarly for the average poses in Figure~\ref{fig:featvis}b).

Next, we look how the linear combination of embedding features encodes the abstract attributes of the person.
We apply PCA on the image embedding vectors of all images in the training set, and then project the image embeddings onto two principal components.
Figure~\ref{fig:featvis}c plots the two PCA coefficients using the same local region visualization as Figure~\ref{fig:featvis}a.  Figure~\ref{fig:featvis}d shows the corresponding plot for the pose embedding.
The first PCA component (x-axis in Figs.~\ref{fig:featvis}c and \ref{fig:featvis}d) encodes the orientation (viewpoint) of the person, while 
the second PCA component (y-axis) encodes the attributes of the legs.  
For example, when the y-value is large the left leg is closer to the camera, while when the y-value is small, the right leg is closer to the camera.

Finally, these visualizations along with the supplemental video show that the learned embedding is smooth, even though the temporal order of frames are not used. We believe this is because the score function is learned using a max-margin constraint, which induces a topology of the embedding. Specifically, since the margin is based on the MPJPE between two poses, then the embedding vectors of any two poses should be at least as far apart (according to inner-product) as their MPJPE.
In addition, the  image and pose embeddings are properly aligned; 97\% of the max-score poses for the training images are within 30 MPJPE of the ground-truth pose.

\section{Conclusion}
\vspace{-0.1in}
In this paper, we propose a maximum-margin structured learning framework with deep neural network for human pose estimation. 
Our framework takes image and pose as inputs and outputs a score value that represents a multi-view similarity between the two inputs (whether they depict  the same pose).
The network consists of a CNN for image feature extraction, and two separate sub-networks for non-linear transformation of the image and pose into a joint embedding, where the dot-product between the embeddings  serves as the score function.
We train the network using a maximum-margin cost function, which enforces a re-scaling margin between the score values of the ground-truth image-pose pair and other image-pose pairs.
This specific form of embedding and score function makes inference computationally efficient, by allowing the pose embedding for a candidate set of poses to be calculated off-line.
We evaluate our proposed framework on Human3.6M dataset and achieve significant improvement over the state-of-art.
Finally, we show that the learned image-pose embedding encodes semantic attributes of the pose, such as the orientation of the person and the position of the legs.
Our proposed framework is general, and future work will consider applying it to other structured-output tasks.

{\small
\bibliographystyle{ieee}
\bibliography{iccv2015_arxiv_v0}
} 
 
\end{document}